\documentclass{article} % For LaTeX2e
\usepackage{nips15submit_e,times}
\usepackage{algorithm}
\usepackage{algpseudocode}
\providecommand{\inlinecode}[1]{\texttt{#1}}

\title{rnn : Recurrent Library for Torch7}

\author{
Nicholas L\'{e}onard \\
Element Inc.\\
New York, NY \\
\texttt{nick@nikopia.org} \\
\And
Sagar Waghmare \\
Element Inc.\\
New York, NY \\
\texttt{sw@discoverelement.com} \\
\And
Yang Wang \\
Element Inc.\\
New York, NY \\
\texttt{yw@discoverelement.com} \\
\And
Jin-Hwa Kim \\
Seoul National University \\
Seoul, Republic of Korea \\
\texttt{jnhwkim@snu.ac.kr} \\
}

\nipsfinalcopy % Uncomment for camera-ready version

\begin{document}

\maketitle

\begin{abstract}
The \textbf{rnn} package provides components for 
implementing a wide range of Recurrent Neural Networks.
It is built withing the framework of the Torch distribution
for use with the \textbf{nn} package.
The components have evolved from 3 iterations, each 
adding to the flexibility and capability of the package.
All component modules inherit either the \inlinecode{AbstractRecurrent}
or \inlinecode{AbstractSequencer} classes.
Strong unit testing, continued backwards compatibility and access to supporting material
are the principles followed during its development.
The package is compared against existing implementations of two published papers.
\end{abstract}

\section{Introduction}
In recent years, deep learning research has seen a resurgence of interest 
in Recurrent Neural Networks (RNN). In the scope of our own research, we
have developed a package that makes it easy to implement a wide range of RNNs using the 
Torch distribution. 
The \textbf{rnn} package\footnote{https://github.com/Element-Research/rnn} 
can be used to implement recurrent neural networks 
like simple RNNs and Long Short Term Memory (LSTM) networks.
The package is very general and makes heavy use 
of object-oriented programming to keep it as simple to use and extend as possible.
The sections are divided into an overview of the \textbf{Torch 7} distribution, 
package components organized historically and principles during its development.

\section{Torch}

Torch\footnote{http://torch.ch/} is a scientific computing distribution 
with a focus on deep learning research and applications \cite{collobert2011torch7}.
The main interface is accessible through the Lua programming language \cite{ierusalimschy1996lua},
which uses functions and structures implemented using the C and CUDA programming languages.
Lua is simple enough to make it easy to implement code for fast execution in C/CUDA. 
Torch 7 has fast and efficient support for Graphical Processing Unit (GPU)
via the \textbf{cutorch} and \textbf{cunn} packages. 
The distribution is used by Facebook, Google DeepMind, Twitter, New York University and 
many other organizations. Through GitHub\footnote{https://github.com/}, one can access documentation, tutorials and 
a wide variety of examples. The project is quite mature as it has been under active
development since October 2012. The distribution is divided into different packages 
which we will overview in the next sections.

\subsection{torch7}

This package is the core of the distribution\footnote{https://github.com/torch/torch7}. 
It provides a \inlinecode{Tensor} class for manipulating multi-dimensional arrays.
This is the main class of objects used in Torch 7.
The \inlinecode{Tensor} supports common operations like Basic Linear Algebra Sub-routines (BLAS), 
random initialization, indexing, slicing, transposition, etc.
Most operations for\inlinecode{FloatTensor} and \inlinecode{DoubleTensor} 
are also implemented for \inlinecode{CudaTensors} (via the \textbf{cutorch}).

While Lua can be used to implement class hierarchies, 
or more generally, object-oriented programming (OOP),
the torch package provides utilities such as \inlinecode{torch.class} for implementing inheritance 
and \inlinecode{torch.serialize} for serialization. The \textbf{torch} package 
also provides utilities for saving objects to disk, unit testing, etc.

\subsection{nn}

This package implements feed-forward neural networks\footnote{https://github.com/torch/nn}.
These form a computational flow-graph of transformation.
They typically learn through backpropagation, 
which is gradient descent using the chain rule \cite{rumelhart2002learning}.

The \textbf{nn} package is very simple as all classes inherit one of either two abstract classes :

\begin{itemize} 
	\item Module : differentiable transformations of \inlinecode{input} to \inlinecode{output} ;
    \item Criterion : cost function to minimize. Outputs a scalar loss;
\end{itemize}

The \textbf{nn} is used by first building a graph of modules using 
composite (\inlinecode{Container} subclasses) and component modules,
and then training a the resulting neural network on some data.

As an example, a Multi-Layer Perceptron (MLP) with 2 layers of hidden units can be 
assembled as such:

% Generator: GNU source-highlight, by Lorenzo Bettini, http://www.gnu.org/software/src-highlite
\noindent
\mbox{}mlp\ =\ nn.\textbf{Sequential}() \\
\mbox{}\textbf{mlp:add}(nn.\textbf{Convert}(\texttt{'bchw'},\ \texttt{'bf'}))\ \textit{-\/-\ collapse\ 3D\ to\ 1D} \\
\mbox{}\textbf{mlp:add}(nn.\textbf{Linear}(1*28*28,\ 200)) \\
\mbox{}\textbf{mlp:add}(nn.\textbf{Tanh}()) \\
\mbox{}\textbf{mlp:add}(nn.\textbf{Linear}(200,\ 200)) \\
\mbox{}\textbf{mlp:add}(nn.\textbf{Tanh}())\  \\
\mbox{}\textbf{mlp:add}(nn.\textbf{Linear}(200,\ 10)) \\
\mbox{}\textbf{mlp:add}(nn.\textbf{LogSoftMax}())\ \textit{-\/-\ for\ classification\ problems}

In the above example, the \inlinecode{Sequential} is a \inlinecode{Container} subclass.
A call to \inlinecode{output = mlp:forward(input)} will iteratively transform the \inlinecode{input}
one module at a time, in order that these were added to the composite.

To train the \inlinecode{mlp} module on a dataset, 
the Negative Log-Likelihood (NLL) criterion could be used:

% Generator: GNU source-highlight, by Lorenzo Bettini, http://www.gnu.org/software/src-highlite
\noindent
\mbox{}nll\ =\ nn.\textbf{ClassNLLCriterion}()

The actual training loop would usually be a variation of the following :

% Generator: GNU source-highlight, by Lorenzo Bettini, http://www.gnu.org/software/src-highlite
\noindent
\mbox{}\textbf{function}\ \textbf{trainEpoch}(module,\ criterion,\ inputs,\ targets) \\
\mbox{}\ \ \ \textbf{for}\ i=1,\textbf{inputs:size}(1)\ \textbf{do} \\
\mbox{}\ \ \ \ \ \ \textbf{local}\ idx\ =\ math.\textbf{random}(1,\textbf{inputs:size}(1)) \\
\mbox{}\ \ \ \ \ \ \textbf{local}\ input,\ target\ =\ inputs[idx],\ \textbf{targets:narrow}(1,idx,1) \\
\mbox{}\ \ \ \ \ \ \textit{-\/-\ forward} \\
\mbox{}\ \ \ \ \ \ \textbf{local}\ output\ =\ \textbf{module:forward}(input) \\
\mbox{}\ \ \ \ \ \ \textbf{local}\ loss\ =\ \textbf{criterion:forward}(output,\ target) \\
\mbox{}\ \ \ \ \ \ \textit{-\/-\ backward} \\
\mbox{}\ \ \ \ \ \ \textbf{local}\ gradOutput\ =\ \textbf{criterion:backward}(output,\ target) \\
\mbox{}\ \ \ \ \ \ \textbf{module:zeroGradParameters}() \\
\mbox{}\ \ \ \ \ \ \textbf{local}\ gradInput\ =\ \textbf{module:backward}(input,\ gradOutput) \\
\mbox{}\ \ \ \ \ \ \textit{-\/-\ update} \\
\mbox{}\ \ \ \ \ \ \textbf{module:updateParameters}(0.1)\ \textit{-\/-\ W\ =\ W\ -\ 0.1*dL/dW} \\
\mbox{}\ \ \ \textbf{end} \\
\mbox{}\textbf{end}

The above \inlinecode{trainEpoch} function could be used to train the
\inlinecode{mlp} module using the \inlinecode{nll} criterion
to fit a classification dataset defined by the \inlinecode{inputs} and 
\inlinecode{targets} tensors. 

The \textbf{rnn} package was designed to be used in the scope of the \textbf{nn} package.
This means that its components must conform to the \inlinecode{Module} and 
\inlinecode{Criterion} interfaces such that these can be used in 
for training with functions like \inlinecode{trainEpoch}.

\section{Package Components}

This section is a kind of analysis of the package, exploring 
its historical development and the components that evolved from it. 
While it would be nice to come up 
with the finished product in the first iteration, often times we only get to such a state 
as time progresses. And in our necessity to maintain a certain level of backwards compatibility,
the final product can only really be understood through its historical development.
As such, we have divided the analysis of its components into the 3 major iterations in 
which they appeared.
 
Before this package, the only way to implement RNNs for variable length sequences was to manually 
clone the recurrent modules for each time-step, have these share parameters and
write code to apply these clones over a sequence. This was against the base
philosophy of the \textbf{nn} package where every transformation of \inlinecode{input} to \inlinecode{output} should 
implemented as a \inlinecode{Module} (or composite thereof). 
So the \textbf{rnn} package started out as a single \inlinecode{Recurrent} module that 
internally implemented a general interface for implementing variations of Simple RNNs
as described in \cite[section 2.5-2.8]{sutskever2013training}, 
\cite[section~3.2-3.3]{mikolov2012statistical} and \cite{boden2001guide}. 
More generally, a recurrent module is responsible for managing the cloning, 
parameter sharing of the and sequentially applying these internal modules
to elements of a sequence. 

\subsection{First Iteration : Recurrent module}

As a first iteration, we wanted to be able to forward a sequence through a \inlinecode{Recurrent}
module by making successive calls to its \inlinecode{forward} method\footnote{For reasons of backwards 
compatibility this use case is still supported} :

% Generator: GNU source-highlight, by Lorenzo Bettini, http://www.gnu.org/software/src-highlite
\noindent
\mbox{}\textit{-\/-\ generate\ some\ dummy\ inputs\ and\ gradOutputs\ sequences} \\
\mbox{}inputs,\ gradOutputs\ =\ \{\},\ \{\} \\
\mbox{}\textbf{for}\ step=1,rho\ \textbf{do} \\
\mbox{}\ \ \ inputs[step]\ =\ torch.\textbf{randn}(batchSize,inputSize) \\
\mbox{}\ \ \ gradOutputs[step]\ =\ torch.\textbf{randn}(batchSize,inputSize) \\
\mbox{}\textbf{end} \\
\mbox{} \\
\mbox{}\textit{-\/-\ an\ AbstractRecurrent\ instance} \\
\mbox{}rnn\ =\ nn.\textbf{Recurrent}( \\
\mbox{}\ \ \ hiddenSize,\ \textit{-\/-\ size\ of\ the\ input\ layer} \\
\mbox{}\ \ \ nn.\textbf{Linear}(inputSize,outputSize),\ \textit{-\/-\ input\ layer} \\
\mbox{}\ \ \ nn.\textbf{Linear}(outputSize,\ outputSize),\ \textit{-\/-\ recurrent\ layer} \\
\mbox{}\ \ \ nn.\textbf{Sigmoid}(),\ \textit{-\/-\ transfer\ function} \\
\mbox{}\ \ \ rho\ \textit{-\/-\ maximum\ number\ of\ time-steps\ for\ BPTT} \\
\mbox{}) \\
\mbox{} \\
\mbox{}\textit{-\/-\ feed-forward\ and\ backpropagate\ through\ time\ like\ this\ :} \\
\mbox{}\textbf{for}\ step=1,rho\ \textbf{do} \\
\mbox{}\ \ \ \textbf{rnn:forward}(inputs[step]) \\
\mbox{}\ \ \ \textbf{rnn:backward}(inputs[step],\ gradOutputs[step]) \\
\mbox{}\textbf{end} \\
\mbox{}\textbf{rnn:backwardThroughTime}()\ \textit{-\/-\ call\ backward\ on\ the\ internal\ modules} \\
\mbox{}gradInputs\ =\ rnn.gradInputs \\
\mbox{}\textbf{rnn:updateParameters}(0.1) \\
\mbox{}\textbf{rnn:forget}()\ \textit{-\/-\ resets\ the\ time-step\ counter}

As can be seen by the above example, the original design allowed for 
the call to \inlinecode{forward} of each element in the sequence to 
be immediately followed by a commensurate call to \inlinecode{backward}.
Since backpropagation through time (BPTT)\cite{rumelhart2002learning} can only occur after the entire 
sequence had been forwarded through the RNN, the above calls to \inlinecode{backward}
cannot perform BPTT. Instead they only keep a copy of the provided \inlinecode{gradOutput} for each time-step.
The actual BPTT required an explicit call to the \inlinecode{backwardThroughTime} 
of all \inlinecode{AbstractRecurrent} instances. 

This design also prevented calls to \inlinecode{backward} from
returning a valid \inlinecode{gradInput}, as these are only made available after BPTT.
This is also what necessitated the second argument of the \inlinecode{Recurrent} 
constructor, which offers a means for handling previous layers internally.

\subsection{Second Iteration : Sequencer and LSTM}

When the \inlinecode{LSTM} module was being implemented during out second iteration, 
it quickly became apparent that constraints resulting 
from our design of the \inlinecode{AbstractRecurrent} were too limiting.
For one, the first iteration made it impossible to stack \inlinecode{AbstractRecurrent} instances.
However, as is often the case with the \textbf{nn} package, the problem 
could be resolved by abstracting these intricacies away into another module.
Hence the \inlinecode{Sequencer} was born.

\subsubsection{Sequencer}
The \inlinecode{Sequencer} module is a decorator used to abstract 
away the intricacies of \inlinecode{AbstractRecurrent} modules
like \inlinecode{Recurrence}, \inlinecode{Recurrent} and \inlinecode{LSTM}. 

% Generator: GNU source-highlight, by Lorenzo Bettini, http://www.gnu.org/software/src-highlite
\noindent
\mbox{}seq\ =\ nn.\textbf{Sequencer}(module)

While an \inlinecode{AbstractRecurrent} instance requires a sequence to be presented one element at a time, 
each with its own call to \inlinecode{forward} (and \inlinecode{backward}), 
the \inlinecode{Sequencer} forwards an entire \inlinecode{input} sequence (a table) 
to yield the resulting \inlinecode{output} sequence (a table of the same length). 
It also takes care of calling \inlinecode{forget}, \inlinecode{backwardOnline} and 
other such \inlinecode{AbstractRecurrent} specific methods.

For example, \inlinecode{rnn}, an \inlinecode{AbstractRecurrent} instance, 
can forward an input sequence one \inlinecode{forward} call at a time:

% Generator: GNU source-highlight, by Lorenzo Bettini, http://www.gnu.org/software/src-highlite
\noindent
\mbox{}input\ =\ \{torch.\textbf{randn}(3,4),\ torch.\textbf{randn}(3,4),\ torch.\textbf{randn}(3,4)\} \\
\mbox{}\textbf{rnn:forward}(input[1]) \\
\mbox{}\textbf{rnn:forward}(input[2]) \\
\mbox{}\textbf{rnn:forward}(input[3])

Equivalently, we can use a \inlinecode{Sequencer} to forward the entire \inlinecode{input} sequence at once:

% Generator: GNU source-highlight, by Lorenzo Bettini, http://www.gnu.org/software/src-highlite
\noindent
\mbox{}seq\ =\ nn.\textbf{Sequencer}(rnn) \\
\mbox{}\textbf{seq:forward}(input)

Furthermore, the \inlinecode{Sequencer} manages the \inlinecode{backward} and 
\inlinecode{backwardThroughTime} calls to the decorated module internally. 
This means that a call to \inlinecode{Sequencer:backward} will return the appropriate
\inlinecode{gradInput} table.

The \inlinecode{Sequencer} can also take a \textit{non-recurrent module}
\footnote{By non-recurrent module, we mean a module that isn't an instance of \inlinecode{AbstractRecurrent},
and that neither contains such instances.}
and apply it to each element of the \inlinecode{input} sequence to produce an \inlinecode{output} table of the same length. 
However, in this second iteration of the package, each \inlinecode{Sequencer} 
instance could only either decorate a recurrent instance
\footnote{Any \inlinecode{AbstractRecurrent} instance is a recurrent instance.},
or a non-recurrent instance.
This was not an imposing constraint as it can be subverted by stacking \inlinecode{Sequencer} instances:

% Generator: GNU source-highlight, by Lorenzo Bettini, http://www.gnu.org/software/src-highlite
\noindent
\mbox{}rnn\ =\ nn.\textbf{Sequential}() \\
\mbox{}\ \ \ :\textbf{add}(nn.\textbf{Sequencer}(nn.\textbf{Linear}(inputSize,\ hiddenSize))) \\
\mbox{}\ \ \ :\textbf{add}(nn.\textbf{Sequencer}(nn.\textbf{LSTM}(hiddenSize,\ hiddenSize))) \\
\mbox{}\ \ \ :\textbf{add}(nn.\textbf{Sequencer}(nn.\textbf{LSTM}(hiddenSize,\ hiddenSize))) \\
\mbox{}\ \ \ :\textbf{add}(nn.\textbf{Sequencer}(nn.\textbf{Linear}(hiddenSize,\ outputSize))) \\
\mbox{}\ \ \ :\textbf{add}(nn.\textbf{Sequencer}(nn.\textbf{LogSoftMax}()))

The above was actually the use-case that brought us to this second iteration of the code base. 
The objective was to build the stacked networks of LSTM layers outlined in \cite{zaremba2014recurrent}.

\subsubsection{LSTM}

The \inlinecode{LSTM} module is an implementation of a layer of 
Long-Short Term Memory units\cite{hochreiter1997long}. 
We used the LSTM in \cite{graves2013speech} as a blueprint for this module as it was the most concise. 
Yet it is also the vanilla LSTM described in \cite{greff2015lstm}.

% Generator: GNU source-highlight, by Lorenzo Bettini, http://www.gnu.org/software/src-highlite
\noindent
\mbox{}module\ =\ nn.\textbf{LSTM}(inputSize,\ outputSize,\ [rho])

The implementation of the \inlinecode{forward} method corresponds to the following algorithm:

\begin{algorithm}
\caption{~~Long Short Term Memory feed forward}
\label{alg:lstm}
\begin{algorithmic}[1]
\State $i_t = \sigma(W_{x\rightarrow i}x_t + W_{h\rightarrow i}h_{t-1} + W_{c \rightarrow i}c_{t-1} + b_{1 \rightarrow i})$
\State $f_t = \sigma(W_{x\rightarrow f}x_t + W_{h\rightarrow f}h_{t-1} + W_{c\rightarrow f}c_{t-1} + b_{1\rightarrow f})$
\State $z_t = \tanh(W_{x\rightarrow c}x_t + W_{h\rightarrow c}h_{t−1} + b_{1\rightarrow c})$
\State $c_t = f_t c_{t-1} + i_t z_t$
\State $o_t = \sigma(W_{x\rightarrow o}x_t + W_{h\rightarrow o}h_{t−1} + W_{c\rightarrow o}c_t + b_{1\rightarrow o})$
\State $h_t = o_t\tanh(c_t)$
\end{algorithmic}
\end{algorithm}

where $W_{s\rightarrow q}$ is the weight matrix from $s$ to $q$, $t$ indexes the time-step, 
$b_{1\rightarrow q}$ are the biases leading into $q$, $\sigma()$ is the logistic function, 
$x_t$ is the input, $i_t$ is the input gate (line 1), $f_t$ is the forget gate (line 2), 
$z_t$ is the input to the cell (which we call the hidden) (line 3), $c_t$ is the cell (line 4), 
$o_t$ is the output gate (line 5), and $h_t$ is the output of this module (line 6). 
Also note that the weight matrices from cell to gate vectors are diagonal $W_{c\rightarrow s}$, 
where $s$ is gate $i$, $f$, or $o$.

The \inlinecode{LSTM} module is implemented internally as a composite of existing modules.
As in the case of the \inlinecode{Recurrent} class, a different clone sharing parameters with the internal module 
is applied to each time-step. Each clone manages its own copy of intermediate representations, 
which consists mostly of \inlinecode{output} and \inlinecode{gradInput} attributes.

\subsubsection{Repeater}

% Generator: GNU source-highlight, by Lorenzo Bettini, http://www.gnu.org/software/src-highlite
\noindent
\mbox{}r\ =\ nn.\textbf{Repeater}(module,\ nStep)

While the \inlinecode{Sequencer} applied a decorated module to an \inlinecode{input} sequence (a table), 
the \inlinecode{Repeater} repeatedly applies a module to a single unchanging \inlinecode{input}. 
Both decorators produce an \inlinecode{output} sequence (a table). 
The \inlinecode{Repeater} was designed to implement things that are recursively applied to the same input,
like Recurrent Convolutional Neural Networks (RCNN)\cite{pinheiro2013recurrent}.

The second iteration arose out of the necessity to allow for stacking of recurrent instances, 
specifically \inlinecode{LSTM} modules. 

\subsection{Third Iteration}

The current iteration arose from the reproduction of the Recurrent Attention Model (RAM) 
described in \cite{mnih2014recurrent}.
The only lacking component to reproduce the RAM was the \inlinecode{RecurrentAttention}
module.

\subsubsection{RecurrentAttention}
This module is similar to the \inlinecode{Repeater} module in that it recursively applies 
an \inlinecode{rnn} module to a fixed \inlinecode{input}, which in this case is an image.

% Generator: GNU source-highlight, by Lorenzo Bettini, http://www.gnu.org/software/src-highlite
\noindent
\mbox{}ram\ =\ nn.\textbf{RecurrentAttention}(rnn,\ action,\ nStep,\ hiddenSize)

The \inlinecode{rnn} argument is an AbstractRecurrent instance
which expects a table \inlinecode{\{x, z\}} as \inlinecode{input} 
where \inlinecode{x} is the \inlinecode{ram} \inlinecode{input} and 
\inlinecode{z} is an action sampled from the \inlinecode{action} module. 

The \inlinecode{action} is a Module that learns using the 
REINFORCE learning rule \cite{williams1992simple}.
It samples actions given the previous time-step's \inlinecode{rnn output}. 
The \inlinecode{action} module's outputs are only used internally to guide 
the attention of the \inlinecode{RecurrentAttention} module.

The implementation of \inlinecode{RecurrentAttention} module was a kind of validation of the
separation of functionality between the \inlinecode{AbstractRecurrent} and \inlinecode{AbstractSequencer} classes.
The first defines general components that handles the recursion from \inlinecode{forward} to the next, i.e. one element at a time.
It is an abstract class inherited by \inlinecode{LSTM} and \inlinecode{Recurrent}.
The second defines how the recurrent component is used for specific tasks involving sequences, i.e. one sequence of elements at a time.
It is an abstract class inherited by \inlinecode{Sequencer}, \inlinecode{Repeater} and \inlinecode{RecurrentAttention}
This division of labor happens to be modular enough to allow for implementing most tasks 
without requiring the writing of new code for both types of modules. What we mean by this is that 
research topics will generally explore modifications of either abstract classes, but not both at the same time.

Nevertheless, the \textbf{rnn} library was still lacking the flexibility
to allow for more complex configurations of non-recurrent instances with recurrent instances.
The solution to this problem arose from the observation that \inlinecode{RecurrentAttention} 
expected the \inlinecode{action} constructor argument to be a non-recurrent instance. 
However, to make the \inlinecode{RecurrentAttention} module 
generalize to the later DRAM implementation in (citation), it would need to allow 
composites of both recurrent and non-recurrent instances for the \inlinecode{action} argument. 
Again, the easiest way to make this happen, was to 
implement a new module, in this case the \inlinecode{Recursor}.

\subsubsection{Recursor}
This module decorates another \inlinecode{module} to allow it to be used within an \inlinecode{AbstractSequencer} instance. 
It does this by making the decorated \inlinecode{module} conform to the \inlinecode{AbstractRecurrent} interface, 
which like the \inlinecode{LSTM} and \inlinecode{Recurrent} classes, this class inherits.

% Generator: GNU source-highlight, by Lorenzo Bettini, http://www.gnu.org/software/src-highlite
\noindent
\mbox{}rec\ =\ nn.\textbf{Recursor}(module[,\ rho])

For each successive call to \inlinecode{updateOutput} (i.e. \inlinecode{forward}), 
this decorator will call \inlinecode{stepClone} on the decorated \inlinecode{module}. 
So for each time-step, it will forward the commensurate input through a commensurate clone of the \inlinecode{module}. 
As usual, both the clone and original share parameters and gradients w.r.t. parameters.
\footnote{For recurrent modules, the clones and original module are one and the same (i.e. no cloning occurs)}

So in the second iteration, to stack \inlinecode{LSTMs}, two \inlinecode{Sequencers} were required :

% Generator: GNU source-highlight, by Lorenzo Bettini, http://www.gnu.org/software/src-highlite
\noindent
\mbox{}lstm\ =\ nn.\textbf{Sequential}() \\
\mbox{}\ \ \ :\textbf{add}(nn.\textbf{Sequencer}(nn.\textbf{LSTM}(100,100))) \\
\mbox{}\ \ \ :\textbf{add}(nn.\textbf{Sequencer}(nn.\textbf{LSTM}(100,100)))

Using a Recursor, the same model can be assembled with a single \inlinecode{Sequencer} :

% Generator: GNU source-highlight, by Lorenzo Bettini, http://www.gnu.org/software/src-highlite
\noindent
\mbox{}lstm\ =\ nn.\textbf{Sequencer}( \\
\mbox{}\ \ \ nn.\textbf{Recursor}( \\
\mbox{}\ \ \ \ \ \ nn.\textbf{Sequential}() \\
\mbox{}\ \ \ \ \ \ \ \ \ :\textbf{add}(nn.\textbf{LSTM}(100,100)) \\
\mbox{}\ \ \ \ \ \ \ \ \ :\textbf{add}(nn.\textbf{LSTM}(100,100)) \\
\mbox{}\ \ \ \ \ \ ) \\
\mbox{}\ \ \ )

Actually, the \inlinecode{Sequencer} will wrap any non-recurrent module into a \inlinecode{Recursor} automatically.
So the above model can be further simplified :

% Generator: GNU source-highlight, by Lorenzo Bettini, http://www.gnu.org/software/src-highlite
\noindent
\mbox{}lstm\ =\ nn.\textbf{Sequencer}( \\
\mbox{}\ \ \ nn.\textbf{Sequential}() \\
\mbox{}\ \ \ \ \ \ :\textbf{add}(nn.\textbf{LSTM}(100,100)) \\
\mbox{}\ \ \ \ \ \ :\textbf{add}(nn.\textbf{LSTM}(100,100)) \\
\mbox{}\ \ \ )

A non-recurrent instance like \inlinecode{Linear} can also be added between both \inlinecode{LSTMs}. 
In this case, a \inlinecode{Linear} will be cloned (and have its parameters shared) for each time-step, 
while the \inlinecode{LSTMs} will handle cloning internally :

% Generator: GNU source-highlight, by Lorenzo Bettini, http://www.gnu.org/software/src-highlite
\noindent
\mbox{}lstm\ =\ nn.\textbf{Sequencer}( \\
\mbox{}\ \ \ nn.\textbf{Sequential}() \\
\mbox{}\ \ \ \ \ \ :\textbf{add}(nn.\textbf{LSTM}(100,100)) \\
\mbox{}\ \ \ \ \ \ :\textbf{add}(nn.\textbf{Linear}(100,100)) \\
\mbox{}\ \ \ \ \ \ :\textbf{add}(nn.\textbf{LSTM}(100,100)) \\
\mbox{}\ \ \ )

To recapitulate, recurrent instances are expected to manage time-steps internally. 
Non-recurrent instances can be wrapped by a \inlinecode{Recursor} to yield the same behavior.

So the final version of the \inlinecode{AbstractSequencer} subclasses
automatically decorate all non-recurrent instances with a \inlinecode{Recursor}. 
This allows the \inlinecode{RecurrentAttention} module to accept 
any type of \inlinecode{action} module, thus providing the required flexibility to use it to 
implement the DRAM model without any modifications to existing modules.

\subsubsection{Recurrence}

The last module introduced in this third iteration is the \inlinecode{Recurrence} module.
Another \inlinecode{AbstractRecurrent} subclass, this module
is an extremely general container for implementing recurrences that feedback the previous \inlinecode{output} 
alongside the current {input} to the \inlinecode{Recurrence}. 

% Generator: GNU source-highlight, by Lorenzo Bettini, http://www.gnu.org/software/src-highlite
\noindent
\mbox{}rnn\ =\ nn.\textbf{Recurrence}(module,\ outputSize,\ nInputDim,\ [rho])

Unlike the older \inlinecode{Recurrent} module, \inlinecode{Recurrence} only 
requires a single \inlinecode{module} which implements the actual 
recurrence internally.
This \inlinecode{module} should forward an \inlinecode{output} a tensor (or table) for the current time-step (\inlinecode{output(t)}) 
given an \inlinecode{input} table : \inlinecode{\{input(t), output(t-1)\}}.
Using a mix of \inlinecode{Recursor} (say, via \inlinecode{Sequencer}) and \inlinecode{Recurrence}, 
it is possible to implement any a very general set of recurrent neural networks, including LSTMs and Simple RNNs.

For the first step, the \inlinecode{Recurrence} forwards a Tensor (or table thereof) of zeros through 
the recurrent layer (like \inlinecode{LSTM}, unlike \inlinecode{Recurrent}).

As an example, let us combine \inlinecode{Sequencer} and \inlinecode{Recurrence} to build a Simple RNN for language modeling :

% Generator: GNU source-highlight, by Lorenzo Bettini, http://www.gnu.org/software/src-highlite
\noindent
\mbox{}\textit{-\/-\ recurrent\ module} \\
\mbox{}rm\ =\ nn.\textbf{Sequential}() \\
\mbox{}\ \ \ :\textbf{add}(nn.\textbf{ParallelTable}() \\
\mbox{}\ \ \ \ \ \ :\textbf{add}(nn.\textbf{LookupTable}(nIndex,\ hiddenSize)) \\
\mbox{}\ \ \ \ \ \ :\textbf{add}(nn.\textbf{Linear}(hiddenSize,\ hiddenSize))) \\
\mbox{}\ \ \ :\textbf{add}(nn.\textbf{CAddTable}()) \\
\mbox{}\ \ \ :\textbf{add}(nn.\textbf{Sigmoid}()) \\
\mbox{} \\
\mbox{}rnn\ =\ nn.\textbf{Sequencer}( \\
\mbox{}\ \ \ nn.\textbf{Sequential}() \\
\mbox{}\ \ \ \ \ \ :\textbf{add}(nn.\textbf{Recurrence}(rm,\ hiddenSize,\ 1)) \\
\mbox{}\ \ \ \ \ \ :\textbf{add}(nn.\textbf{Linear}(hiddenSize,\ nIndex)) \\
\mbox{}\ \ \ \ \ \ :\textbf{add}(nn.\textbf{LogSoftMax}()) \\
\mbox{})

Both the \inlinecode{input} and \inlinecode{output} of the \inlinecode{rnn} module will 
be a table of tensors. For example :

% Generator: GNU source-highlight, by Lorenzo Bettini, http://www.gnu.org/software/src-highlite
\noindent
\mbox{}input\ =\ \{\} \\
\mbox{}\textbf{for}\ i=1,rho\ \textbf{do} \\
\mbox{}\ \ \ table.\textbf{insert}(input,\ torch.\textbf{Tensor}(batchSize):\textbf{random}(1,nIndex)) \\
\mbox{}\textbf{end} \\
\mbox{}output\ =\ \textbf{rnn:forward}(input) \\
\mbox{}\textbf{assert}(\#output\ ==\ \#input)

RNNs require sequential data. In the above example, the \inlinecode{input} is a sequence 
of \inlinecode{LookupTable} indices. If the task is to predict the next word given 
the previous word(s) (i.e. language modeling), then the \inlinecode{target} would also be 
a sequence of indices.

If however we only wanted to use the \inlinecode{rho} previous time-steps (words) to predict 
a single target word, we could do so by having the output layer depend 
only on the most recent \inlinecode{output(t)} of the \inlinecode{rnn}. 

For example if we want to do sentiment analysis \cite{pang2008opinion}, 
we could use something like the following :

% Generator: GNU source-highlight, by Lorenzo Bettini, http://www.gnu.org/software/src-highlite
\noindent
\mbox{}\textit{-\/-\ recurrent\ module} \\
\mbox{}rm\ =\ nn.\textbf{Sequential}() \\
\mbox{}\ \ \ :\textbf{add}(nn.\textbf{ParallelTable}() \\
\mbox{}\ \ \ \ \ \ :\textbf{add}(nn.\textbf{LookupTable}(nIndex,\ hiddenSize)) \\
\mbox{}\ \ \ \ \ \ :\textbf{add}(nn.\textbf{Linear}(hiddenSize,\ hiddenSize))) \\
\mbox{}\ \ \ :\textbf{add}(nn.\textbf{CAddTable}()) \\
\mbox{}\ \ \ :\textbf{add}(nn.\textbf{Sigmoid}()) \\
\mbox{}\textit{-\/-\ full\ RNN} \\
\mbox{}rnn\ =\ nn.\textbf{Sequential}() \\
\mbox{}\ \ \ :\textbf{add}(nn.\textbf{Sequencer}(nn.\textbf{Recurrence}(rm,\ hiddenSize,\ 1))) \\
\mbox{}\ \ \ :\textbf{add}(nn.\textbf{SelectTable}(-1))\ \textit{-\/-select\ last\ element} \\
\mbox{}\ \ \ :\textbf{add}(nn.\textbf{Linear}(hiddenSize,\ nSentiment)) \\
\mbox{}\ \ \ :\textbf{add}(nn.\textbf{LogSoftMax}()) \\
\mbox{})

\section{Development Principles}

The previous section discussed the main components used in the \textbf{rnn} package, and how they 
evolved from the need for additional functionality or new use cases.
In all these cases, we didn't go into too much details regarding the internal workings of 
each model. For example, we did not discuss the ability of \inlinecode{Sequencers}
to remember previously presented sequences, the ability of recurrent
instances to evaluate very long sequences without requiring any additional memory,
the ability of all modules to deal with nested tables of tensors, the ability to handle variable 
length inputs, or how RNNs can BPTT for less time-steps than the number of forwarded time-steps.

\subsection{Unit Testing}
In any cases, each of these features potentially introduce bugs. The only way to 
make sure that these are weeded out and not introduced in later revisions is by 
emphasizing the requirement for broad unit tests. The \textbf{rnn} has unit tests 
for each of its component modules. It also includes unit tests for different combinations 
of modules. These unit tests are almost always designed the same way. Functionality of 
modules introduced by the package is compared to a baseline which is known to work.
For example, when implementing the \inlinecode{Recurrent} module unit tests, it was 
compared to an equivalent composite structure built using modules taken directly from the \textbf{nn} package (which are already unit tested).

However, unit tests can only go so far. The ultimate test is to reproduce the results of existing papers.
For the \inlinecode{LSTM} module, we were initially unable to reproduce the LSTM paper (citation). 
The implementation of the paper was available on GitHub \footnote{https://github.com/wojzaremba/lstm}, 
and used a combination of the \inlinecode{nngraph} package and 
custom code to implement an stack of LSTMs.
So we ended up extracting the code from the original repository that we wanted to reproduce.
We included it in a unit test that tried to have our \inlinecode{LSTM} module match the behavior 
of their own implementation. It was only in doing so that we were able to resolve hidden discrepancies (bugs).
Noteworthy among them was the fact that their LSTM implementation used the last hidden states of the 
previous sequence to influence the current sequence. This was not obvious for us as doing so for 
a Simple RNN introduced instability which often led to divergence during training.
In any case, this massive unit test now ensures that our \inlinecode{LSTM} implementation matches 
a published open-source state of the art implementation.

\subsection{Backward Compatibility}

Since November 2014, the \textbf{rnn} package has been available for use on GitHub as an BSD-licensed open-source repository.
As can be seen by the above overview of its major iterations, the design has evolved over time.
From the start, we have tried to maintain its backward compatibility so that users 
can continue to benefit from updates without requiring major changes to existing code 
or serialized objects that depends on the \textbf{rnn} package.

However, maintaining backwards compatibility has its drawbacks.
For example, the \inlinecode{Recurrent} module is convoluted compared to the newer 
\inlinecode{Recurrence} module. Users will continue to use the former even though 
the latter is more general and easier to use. 

As for the \inlinecode{LSTM} code, it is basically made redundant by the new \inlinecode{Recurrence} module.
A compromise worthy of consideration is to make the \inlinecode{LSTM}
module a \inlinecode{Recurrence} subclass. But this would break backwards compatibility
for users loading an serialized instance of the older \inlinecode{LSTM} 
in the scope of a version of the \textbf{rnn} including the newer \inlinecode{LSTM} instance.
This issue is caused by the way Torch handles serialization and deserialization of objects.
The class definition (i.e. the Lua \textit{metatable}) is not serialized but is required prior to 
serialization. Therefore to implement this compromise, we would still be breaking backwards compatibility 
for serialized modules. But such a change, if implemented correctly (by preserving the same interface), 
would not break existing scripts making use of the \inlinecode{LSTM} module. 
We opted to preserve the \inlinecode{LSTM} code in its current state.

The constraint for backwards compatibility is an important one as it minimizes the hassle for users. 
But at the same time, it does result in redundant code (multiple ways of doing the same thing)
and support for deprecated use cases.

\subsection{Supporting Material}

From its inception, our focus has been on providing supporting material.
We can divide these into the following categories :

\begin{itemize}
\item Documentation;
\item Examples; and
\item Tutorials.
\end{itemize}

Documentation is provided for all modules and criterions provided in the package. 
It also includes references to related scientific articles, examples and tutorials.
Documentation is used as a kind of reference manual for specific components.
Examples are concrete demonstrations of the capabilities of the package with respect to 
implementing a particular use case. These also demonstrate how the package can 
be used with other packages, or more generally, within the scope of the Torch distribution.
The package references example scripts for training language models and
a recurrent attention models on different datasets.
Tutorials include videos, articles or blog posts explaining how to use the package, 
often with respect to a concrete example.

All the supporting material is important as it brings the package to life, 
allowing the user to learn how to use it. It also has the side-effect of 
making it seem more legitimate, thereby encouraging new users to dive in.

\subsection{Core Extensions}

Submitting a GitHub Pull Request (PR) to get some specific code merged into the core packages 
can be daunting. Delays can range from days to weeks. After which the PR is sometimes refused.
Lua has a certain advantage here over other programming languages as its heavy reliance on
tables makes it very easy to overwrite or extend core package functionality from within an non-core package.

For example, the core implementation of the \inlinecode{Module:type()} would 
decouple share parameters. But because the \inlinecode{Module} class definition 
is just another table, it was easy to overwrite the method to 
preserve sharing semantics when type-casting. Many more such core extensions were 
necessary to make the \textbf{rnn} package.

\section{Results}

The package was used to reproduce two papers :
Recurrent Neural Network Regularization \cite{zaremba2014recurrent}
and Recurrent Models for Visual Attention \cite{mnih2014recurrent}.

The first paper implements a stack of LSTM layers \cite{hinton2012improving}
and benchmarks various sizes of the model on different datasets.
The results presented in the paper are better than 
those that can be attained using their commensurate GitHub repository 
\footnote{https://github.com/wojzaremba/lstm}. 
The provided code allows one to train a stack of LSTM layers, 
with and without dropout, 
on the Penn Tree Bank dataset \cite{marcus1993building}.
Using their script, test set perplexity
with and without dropout is 82 and 115, respectively.
Using the \textbf{rnn} package, our script was able to reach 
commensurate perplexities of 83 and 115\footnote{https://github.com/nicholas-leonard/dp/blob/master/examples/recurrentlanguagemodel.lua}.

The second paper implements recurrent attention model (RAM) that 
learns using a combination of backpropagation and REINFORCE\cite{williams1992simple} learning.
The authors do not provide code, but their paper includes a detailed 
description of the model.
The RAM was implemented using the \inlinecode{Recurrent} and \inlinecode{RecurrentAttention}
modules of the \textbf{rnn} package 
\footnote{https://github.com/Element-Research/rnn/blob/master/examples/recurrent-visual-attention.lua}
As specified in the original paper, the RAM is trained on the MNIST \cite{lecun1998mnist} and Translated MNIST datasets.
While they respectively reach 1.07\% and 1.22\% error on both datasets, 
our implementation was able to surpass these results by reaching 0.85\% and 1.14\% error.

\section{Conclusion}

In this paper, we discussed the evolution of the \textbf{rnn} package, its different component
modules, the various principles underlying its development, and its performance
compared to empirical results of published RNN models.

Unlike other RNN implementations using Torch, the \textbf{rnn} package 
doesn't depend on the \textbf{nngraph} library.
Like the \textbf{nn} package, this one is designed with the assumption that all
transformations and loss functions can be refactored 
into either a \inlinecode{Module} or a \inlinecode{Criterion}, respectively.
It can also be used with the official \textbf{optim} 
or the unofficial \textbf{dp} numeric optimization packages.

\bibliography{rnn_library}
\bibliographystyle{abbrv}

\end{document}